\documentclass[journal]{IEEEtran} 
\IEEEoverridecommandlockouts
\pdfoutput=1
\usepackage{amsmath,amssymb,amsfonts}
\usepackage{algorithmic}
\usepackage{graphicx}
\usepackage{textcomp}
\usepackage{xcolor}
\usepackage{caption}
\usepackage{biblatex}  
\usepackage{booktabs}

\def\BibTeX{{\rm B\kern-.05em{\sc i\kern-.025em b}\kern-.08em
    T\kern-.1667em\lower.7ex\hbox{E}\kern-.125emX}}
\begin{document}

\title{Enhance-NeRF: Multiple Performance Evaluation for Neural Radiance Fields\\
}

\author{
    \IEEEauthorblockN{Qianqiu Tan\textsuperscript{1}}
    , \IEEEauthorblockN{Tao Liu\textsuperscript{1}}
    , \IEEEauthorblockN{Yinling  Xie\textsuperscript{2}}
    , \IEEEauthorblockN{Shuwan Yu\textsuperscript{1}}
    , \IEEEauthorblockN{Baohua Zhang\textsuperscript{1 *}}
    \\
    \IEEEauthorblockA{\textsuperscript{1}NJAU},
    \IEEEauthorblockA{\textsuperscript{2}NJCIT}
   
}

\twocolumn[{
\renewcommand\twocolumn[1][]{#1}
\maketitle
\begin{center}
    \includegraphics[width=0.95\textwidth]{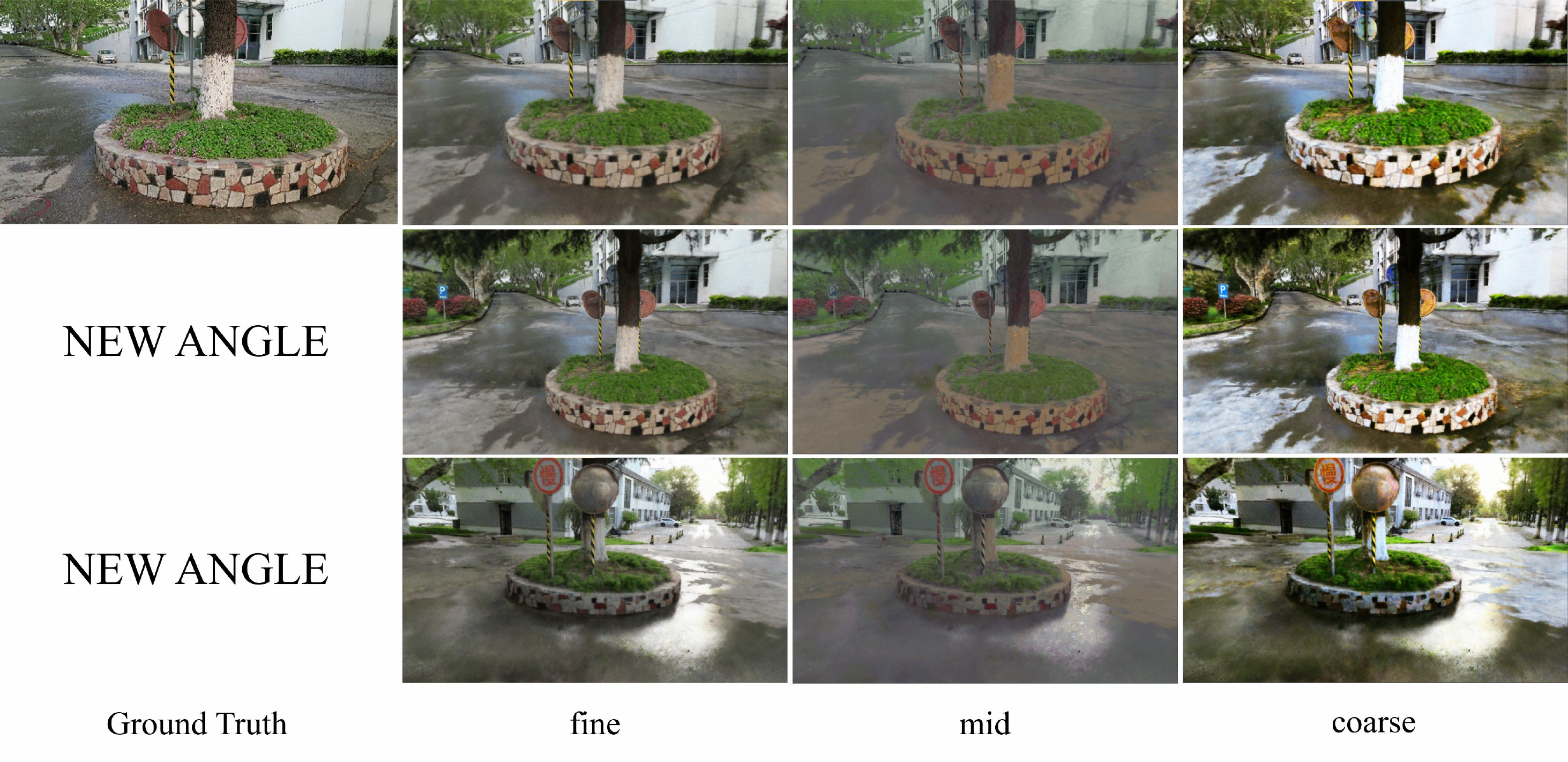}
\end{center}
    \captionof{figure}{\textbf{Multiple Performance Rendering Results.} From Left to Right: Ground Truth, Joint Color (Fine), View-dependent Color (Mid), View-independent Color (Coarse). Our Enhance-NeRF Separates Light Reflection and Non-reflection Areas, and the Synthesis of New Perspectives can Effectively Suppress Color Haze Interference by Supervising Concentration.}
}]

\section*{Abstract\vspace{-1em}}
The quality of three-dimensional reconstruction is a key factor affecting the effectiveness of its application in areas such as virtual reality (VR) and augmented reality (AR) technologies. Neural Radiance Fields (NeRF) can generate realistic images from any viewpoint. It simultaneously reconstructs the shape, lighting, and materials of objects, and without surface defects, which breaks down the barrier between virtuality and reality. The potential spatial correspondences displayed by NeRF between reconstructed scenes and real-world scenes offer a wide range of practical applications possibilities. Despite significant progress in 3D reconstruction since NeRF were introduced, there remains considerable room for exploration and experimentation. NeRF-based models are susceptible to interference issues caused by colored "fog" noise. Additionally, they frequently encounter instabilities and failures while attempting to reconstruct unbounded scenes. Moreover, the model takes a significant amount of time to converge, making it even more challenging to use in such scenarios. Our approach, coined Enhance-NeRF, which adopts joint color to balance low and high reflectivity objects display, utilizes a decoding architecture with prior knowledge to improve recognition, and employs multi-layer performance evaluation mechanisms to enhance learning capacity. It achieves reconstruction of outdoor scenes within one hour under single-card condition. Based on experimental results, Enhance-NeRF partially enhances fitness capability and provides some support to outdoor scene reconstruction. The Enhance-NeRF method can be used as a plug-and-play component, making it easy to integrate with other NeRF-based models. The code is available at: https://github.com/TANQIanQ/Enhance-NeRF

\section{Introduction}
With the concept of Metaverse being proposed, VR and AR technologies are becoming more and more important. Three-dimensional reconstruction is a crucial process for these two technologies, which uses computer technology to transform two-dimensional images or point cloud data into three-dimensional models. It has been widely applied in various fields, such as geological exploration, medical image processing, animated film production, autopilot, and game development. The purpose of representing objects or scenes in a three-dimensional form is to enhance the perceptual experience of humans or the perceptual ability of robots. At present, with the continuous improvement of hardware technology and related algorithms, both the precision and efficiency of three-dimensional reconstruction have been significantly improved.
\par
Three-dimensional dense reconstruction can be achieved through various methods, among which the commonly used ones include RGBD-based \cite{izadi2011kinectfusion}, Simultaneous Localization And Mapping (SLAM) \cite{whelan2016elasticfusion}, Multi-View Stereo (MVS) \cite{yao2018mvsnet}, and Neural Radiance Fields (NeRF) \cite{mildenhall2021nerf}. However, except for NeRF, these methods require depth information to generate results (SLAM and MVS can predict depth information and construct dense point clouds). In this scenario, these techniques are plagued by a critical deficiency——the depths of unobserved space cannot be extrapolated and inferred, as the depth values of the points or voxels can only be obtained from the input images. This predicament is partly attributed to the fact that an image solely provides a restricted view that overlooks other objects that may occupy different locations. As such, the reconstructed scenes generated by these methods suffer from severe surface artifacts. At the same time, depth values have significant errors in outdoor environments. It is difficult to obtain or project accurate depth values for objects at far distances in unbounded views.
\par
In the past two years, NeRFs have received widespread attention as a new technique for 3D representation. NeRF has an amazing performance in 3D representation without surface defect problems. It can perform well in non-occluded regions and unseen scenes. By just giving pictures input with coordinates and a multilayer perceptron (MLP), it outputs the corresponding volume density and color. After volume rendering, any 3D coordinate and angle of view can generate a clear picture.
\par
On the one hand, NeRF exhibits impressive performance across different task sets. By dividing the senses into static, uncertain, and transient parts, NeRF-W \cite{martin2021nerf} can reconstruct sensory inputs using photos from the Internet. Using similar model settings, static-dynamic decomposition tasks can also be addressed effectively \cite{chen2022flow}. On the other hand, NeRF can achieve enhanced capabilities through integration with other methods. By leveraging diffusion models, NeRF can achieve robust results in various 3D-aware image synthesis tasks, including unconditional generation and single/sparse-view 3D reconstruction \cite{chen2023single}. By leveraging clip models, LERF \cite{kerr2023lerf} can extract 3D relevancy maps for a broad range of language prompts interactively in real-time.
\par
Due to the design of view-dependence and volume rendering method, NeRF is able to learn both lighting and material simultaneously, unlike traditional reconstruction methods that can only achieve shape reconstruction and rely on post-processing texture mapping. By jointly reconstructing radiance fields and estimating a physically-based model,  TensoIR \cite{jin2023tensoir} achieves photo-realistic novel view synthesis.
\par
NeRF has shown its potential in autopilot \cite{tancik2022block,rematas2022urban,li2022read}, motion capture \cite{geng2023learning,liu2023hosnerf} and virtual reality. Unfortunately, NeRF still deal with some challenges, such as the optimization of large-scale scenes and the landing application of landing application.

\begin{itemize}
    \item View-dependence is the key feature that enables NeRFs to display lighting changes from different angles \cite{verbin2022ref}. However, many physical surfaces have a low degree of view-dependence when represented in 3D, such as walls, dry ground, and air. The role of view-dependence in color generation may result in "foggy" appearances with irregular colors within the scene \cite{warburg2023nerfbusters}.
    \item Each optimization step in the MLP affects all parameters, meaning rays that are input later will overwrite weights learned from previous input \cite{zhu2022nice}. If an undesirable event occurs, such as poor light sampling, it can significantly decrease the model's ability to remember the scene. This can lead to loss of details when reconstructing the model in more complex and larger scale scenarios.
\end{itemize}

\par
It has been proposed that NeRFs can achieve better performance in rendering clear surfaces by incorporating additional inputs such as depth supervision \cite{deng2022depth}. Additionally, using the Signed Distance Function (SDF) instead of volumetric rendering has also demonstrated favorable results \cite{wang2021neus,yariv2021volume,yu2022monosdf}.By utilizing more demanding data collection process or training techniques that require higher cost, they have successfully attained desirable outcomes. In this paper, our aim is to optimize the potential for reconstructing RGB images so that our model can be applied to a wider range of input scenarios, specifically by reducing the impact of non-overlapping regions and light reflections, which may cause the reconstructed scene to be covered by fog. Resolving them together is difficult because these issues are usually addressed by optimizing the "front-end" processing, such as optimizing the image acquisition scheme and camera pose estimation. \cite{meuleman2023progressively,li2022read,wu2022scalable}. However, due to the computational limitations, we hope to alleviate these challenges by exploring ways to enhance the neural radiation field, to obtain a pure NeRF-based solution. Make certain optimizations for NeRF's 3D object detection \cite{geng2023learning}, semantic segmentation and other tasks.
\par
\textbf{Enhance-NeRF} is heavily influenced by many papers, such as Mip-NeRF 360\cite{barron2022mip}, Instant-NGP \cite{muller2022instant,li2022nerfacc}, Nerfacto \cite{tancik2023nerfstudio}, Nerf-W \cite{martin2021nerf}. Enhance-NeRF combines their features and enables itself to converge quickly with the use of only a small fused MLP. There are two parts insert into our method. First, divide color into view-dependent color(mid) and view-independent color(coarse) as shown in \textbf{Fig. 1}. View-dependent color get the highlight and reflections part in the scenes, while the view-independent color has more interest in the lowlight part. Second, we use different lever SHEcoding to evaluate the three colors separately, guiding our method to enhance the learning of highlight part. 
\par
In summary, our contributions include the following:

\begin{itemize}
    \item We propose a method of joint color, add view-independent color in model training and rendering. Multiple estimations were made for the same pose input without increasing the sampling points of the sampling network.
    \item We propose a method of multiple evaluations using SHEcoding, which improves the utilization of RGB image information and guides the model to learn desired features. 
    \item We have developed a compact model that is capable of efficiently generating highly detailed outdoor scenes using NeRF under multiple image acquisition formats, all within an hour and without imposing significant GPU memory costs.
\end{itemize}

\section{Related works}

\noindent\textbf{\textit{Neural Radiance Fields. }}Vanilla NeRF employs volumetric rendering techniques to represent objects or scenes as three-dimensional spaces composed of particles, simulating complex lighting effects through the scattering and absorption of particles by rays of light. The NeRF field takes the coordinate point $\mathbf{x}=\left(x,y,z\right)$ and direction $\mathbf{d}=(\emptyset,\varphi)$ as input, and gives the density $\sigma$ as output via a spatial MLP. The output from the spatial MLP, together with the encoding of the direction $\mathbf{d}$, are fed into the directional MLP to obtain the color value. By volume rendering, the color of the ray $\mathbf{r}(t)=\mathbf{o}+\mathrm{td}$ can be calculated with this representation:

\begin{equation}
\mathbf{C}=\sum_{k=1}^{K}T\left(t_k\right)\alpha\left(\sigma\left(t_k\right)\delta_k\right)\mathbf{c}\left(t_k\right)
\end{equation}

\noindent where $\mathrm{T}(t_k) =\exp\left(-\sum_{k^\prime=1}^{k-1}{\sigma(t_k)\delta(t_k)}\right)$, $\delta_p=t_{k+1}-t_k$, and $\alpha(x)=1-\exp(-x)$. Vanilla NeRF employs a "coarse-to-fine" methodology, where the coarse MLP's output serves as weights to direct the fine sample points. The final color is rendered using both the coarse and fine samples. The loss of the Vanilla NeRF is:

\begin{equation}
\mathcal{L} = {\sum\limits_{r \in R}\left\lbrack {{\parallel {\hat{C}}_{c}(r) - C(r) \parallel}_{2}^{2} - {\parallel {\hat{C}}_{f}(r) - C(r) \parallel}_{2}^{2}} \right\rbrack}
\end{equation}

\noindent Enhance-NeRF has the same architecture, which utilizes only the position $\mathbf{x}$ and direction $\mathbf{d}$, and a fused MLP that combines both spatial and directional MLPs. In this paper, we denote NeRF's field as:

\begin{equation}
\left. F_{\Theta}:\left( {\mathbf{x},\mathbf{d}} \right)\rightarrow\left( {c,\sigma} \right) \right.
\end{equation}

\noindent \textbf{\textit{Lager scene synthesis}}. To improve the performance of view synthesis in unbounded scenes and enhance the details, it is possible to differentiate the importance of subjects based on their distance from the camera. This can be achieved by employing space warping techniques. In Vanilla NeRF, forward-facing distant views are rendered using the NDC \cite{mildenhall2021nerf}. Mip-Nerf360 \cite{barron2022mip} employs convergence to enhance rendering of faraway scenes. F2-NeRF \cite{wang2023f2nerf} introduces a method called perspective warping to handle rendering in challenging situations. Enhance-NeRF leverage the scene parameterization similar to Mip-Nerf360. MLP-based NeRF has the wonderful advantage of simple development, no need for CUDA programming, and easy model reproducibility. However, when training the model, all parameters in the MLP layers are updated, which can lead to the phenomenon of forgetting previously learned information: During the update process of larger-scale scenes, subsequent updates will impact the entire already-constructed scene and cause blurriness in its reconstruction. Block-NeRF \cite{tancik2022block,} divides the scene into multiple blocks for separate training, and performs appearance matching on all adjacent blocks to achieve alignment and connection between them. This allows rendering to be scaled up to arbitrarily large environments and enables per-block updates. NICE-SALM \cite{zhu2022nice} freezes multiple pre-trained decoders and jointly updates hierarchical scene representations for different scales of scenes. Also, adding another supervision, NeRFs can deal with the problem of unbounded view, they often require specific types of input such as omnidirectional views \cite{jang2022egocentric} or specialized imagery like proxy geometry \cite{wu2022scalable}, or drone shots. For example, Mega-NeRF \cite{turki2022mega} uses drone shots as input and introduces a geometric clustering algorithm to enable data parallelism, allowing for the parallel training of NeRF submodules. Urban Radiance Fields \cite{rematas2022urban} uses laser scan data and RGB images as inputs. Unlike the previous method, Enhance-NeRF utilizes only RGB data and is developed using a pure MLP scheme with PyTorch, achieving reconstruction of outdoor unbounded scenes within one hour and running on a single graphics card.
\par
\noindent \textbf{\textit{Lighting appearance structured.}} In the field of computer graphics, the rendering of object surface highlights and appearance is usually achieved through precomputation techniques \cite{ramamoorthi2009precomputation}. The bidirectional reflectance distribution function (BRDF) \cite{ramamoorthi2002frequency,rusinkiewicz1998new} is used to describe the material of the object, and the reflected light from the object surface is obtained by spherical convolution \cite{ramamoorthi2001signal} of the incident light with the BRDF. By storing this convolution result as prefiltered environment maps \cite{kautz2000approximation,ramamoorthi2002frequency} in advance, rays intersecting the object can be rendered in real-time. The directional MLP in NeRF plays a similar role to prefiltered environment maps, but NeRF's approach is not based on precomputation. Instead, it directly recovers the rendering of a model of the scene from images. Ref-NeRF \cite{verbin2022ref} proposes that the specular reflections on the object surface in the NeRF method are synthesized by emitting light isotopically from points inside the object in volume rendering, rather than being generated by reflecting incident light rays on the object surface. As a result, specular reflections rendered by NeRF are often translucent. " In Ref-NeRF, reflected light is utilized instead of incident light to generate view-dependent color via directional MLP. The spatial MLP is then used to output both a diffuse color (referred to as 'coarse color' in our paper) and a specular tint. Synthesize view-dependent color and diffuse color with reflected light using specular tint, to form the composite color. The composite color is modulated by a fixed tone mapping function to obtain the color output of Ref-NeRF. We explain the composition of surface reflection from different angles and decomposed the color through view-dependent and view-independent methods to obtain a simplified model that ensures the efficiency of the model.

\begin{figure*}[htbp]
    \centering
    \includegraphics[width=0.95\textwidth]{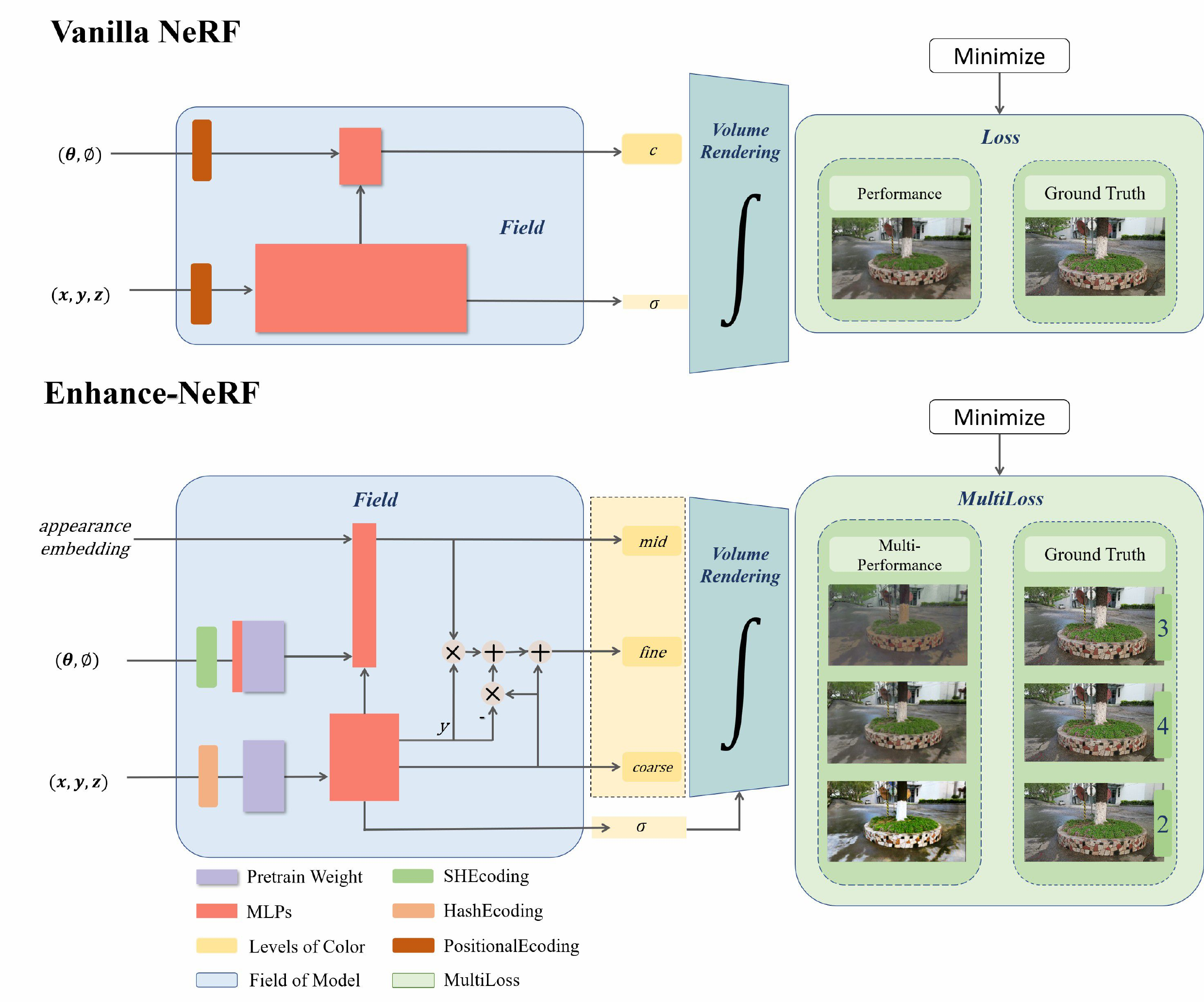}
    \caption{\textbf{Enhance-NeRF}, like Vanilla NeRF, generates $c$ and $\sigma$ through spatial MLP and direction MLP without additional supervision. The added part dealing with viewpoint dependency and conditioning factor y enables modulation of lighting as well as supervision of color in spatial MLP to reduce the effect of atmospheric haze in scenes.}
\end{figure*}

\section{Method}
Enhance-NeRF manages to balance model efficiency, scene reconstruction scale, and performance quality altogether. Our objective is to restore surface reflection effects of objects in unbounded outdoor scenes and alleviate the issue of “fog” noise. Furthermore, we aim to achieve this within an hour using a consumer-grade graphics card and ensuring the model's stability. \textbf{Fig. 2} provides an overview of our method. We selected the Vanilla NeRF method as the baseline and augmented the functionality of NeRF by incorporating joint branches of other commonly used models. Deploying the multi-color combination and multi-level evaluation method of Enhance-NeRF as a plug-and-play component facilitates effortless integration with other NeRF-based models.
\par
To increase the efficiency of building our NeRF model, we utilized Nerfstudio \cite{tancik2023nerfstudio} as our framework, which offers plugins specifically designed for NeRF-based models. This significantly improved the smoothness of our development process. In addition to that, Nerfstudio offers a pipeline consisting of DataManager and DataParsers which can be combined with Colmap to estimate image poses and generate datasets. Moreover, a more versatile representation of 3D space using Frustum has been adopted for both point-based and volume-based samples.

\subsection{Sampling and Scene contraction}
We adopt a “Coarse-to-Fine Online Distillation” structure and the piece-wise sampling strategy from Nerfacto \cite{tancik2023nerfstudio} to map 3D positions to density values. The sampler initially samples uniformly up to a fixed distance from the camera, then gradually increases the step size with each sample to efficiently capture distant objects while maintaining dense sampling of nearby ones. These samples are then fed into a proposal network sampler. The proposal network in Nerfacto features two density fields that iteratively reduce the number of samples, enabling efficient and accurate object sampling even at great distances. We transform these density values into probability distributions along the ray and supervise them to be consistent with NeRF’s density output, to reduce the number of samples taken along each ray during training. The histogram approximation is used as the loss, denoted as $\mathcal{L}_{prop}(\mathbf{t},\mathbf{w},\hat{\mathbf{t}},\hat{\mathbf{w}})$. To accommodate unbounded scenes, the Mip-NeRF360 \cite{barron2022mip} method was introduced to compress space into cubes with a side length of 2. However, in our approach, we use $L^\infty$ norm contraction instead of $L^2$ norm contraction. This modification helps contract space into an aligned cube.

\begin{equation}
contrac{t\left( \mathbf{x} \right)} = \left\{ \begin{matrix}
{x,} \\
{\left( {2 - \frac{1}{\mid \mid \mathbf{x} \mid \mid}} \right)\left( \frac{\mathbf{x}}{\mid \mid \mathbf{x} \mid \mid} \right)} \\
\end{matrix} \right.\begin{matrix}
 & {\mid \mid \mathbf{x} \mid \mid \leq 1} \\
, & {\mid \mid \mathbf{x} \mid \mid > 1} \\
\end{matrix}~
\end{equation}

\subsection{The Field of Enhance-NeRF}
Vanilla NeRF assumes the consistency of scene lighting, but outdoor scenes exhibit characteristics of rapidly changing lighting conditions, with complex light reflection when reconstructing under sunny weather. In order to better adapt the model to different lighting conditions, we adopt NeRF-W \cite{martin2021nerf}, which uses appearance embedding of each image as one of the inputs, providing potential generation guidance for the model. The model can be expressed as \textbf{Equation (5)}, where $c_{fine}$, $c_{mid}$, $c_{coarse}$, $\mathbf{a}$ respectively represent the final color, view-dependent color, view-independent color, and appearance embedding.

\begin{equation}
\left. F_{\Theta}:\left( {\mathbf{x},\mathbf{d},~\mathbf{a}} \right)\rightarrow\left( {\mathbf{C}\left( {c_{fine},c_{mid},c_{coarse}} \right),\sigma} \right) \right.
\end{equation}

Due to the tendency of neural networks towards learning low-frequency signals, encoding of position and direction inputs is required for the model to learn high-frequency information such as texture. Instant-NGP \cite{muller2022instant} utilizes multi-resolution hashing encoding to project position. Enhance-NeRF retains single-resolution hashing encoding to encode the position. We encode direction using spherical harmonics to compose a system that has similar functions to BRDF.
\par
MLP updates globally at each step of training, leading to the forgetting phenomenon that prevents the model from effectively learning and rendering large scenes. Additionally, NeRF’s generalization problem makes it unable to extend to new scenes, causing learned features to be unusable. GRF \cite{trevithick2021grf}  extracts local features of pixels in images and projects them onto 3D points to produce a universal and rich point representation, enabling generalization in NeRF. The bottom-up feature transfer in the era of large image models has inspired a new perspective for NeRF generalization. In PEFT research of image models, AIM\cite{yang2023aim}  locked model parameters and added temporal processing modules, objective functions, and other peripheral modifications to enable image models to perform video understanding tasks without retraining video models. CLIP \cite{radford2021learning}has already shown excellent results even in ZeroShot learning settings. That is, a well-trained image model can extract visual features that are generalizable and effective. However, continued fine-tuning can lead to catastrophic forgetting. If a small amount of data is used for fine-tuning on a large model, it may directly overfit or lose many features of the large model.
\par
In NICE-SLAM \cite{zhu2022nice}, the fixed-weight MLPs decoder are used to avoid forgetting of the network and stabilize updates in large-scale scenes. In Enhance-NeRF, the encoded pose and direction inputs are fed into pre-trained decoders, which include coarse-level and fine-level from NICE-SLAM. Interestingly, the fixed decoder does not appear to significantly improve object shape in rendering results. But it does make the model more stable in reconstruction success rates (almost unaffected by the results brought about by the random initialization of MLPs) and high-dimensional information learn like light intensity. 
\par
Vanilla NeRF generates view-dependent colors for every sampled point in space. Due to the reflection of light and the coordinates only $\sigma$, NeRF’s estimation of concentration $\sigma$ is not accurate. As NeRFs use volume rendering instead of surface rendering, it causes objects to have semi-transparency or a foggy shell \cite{verbin2022ref}. To better distinguish between reflection and non-reflection areas and clean the space, Enhance-NeRF adds a View-independent component to the color estimation process to make the boundary between clear reflection and non-reflection areas more defined. Inspired by momentum, the adjustment form is shown as \textbf{Equation (8)}, and the Field of Enhance-NeRF can be divided into spatial MLP $(F_{\Theta1})$ and direction MLP $(F_{\Theta2})$, just like Vanilla NeRF.

\begin{equation}
\left. F_{\Theta 1}:\left( {\mathbf{d},\mathbf{a}} \right)\rightarrow\left( c_{mid} \right) \right.
\end{equation}

\begin{equation}
\left. F_{\Theta 2}:\left( \mathbf{x} \right)\rightarrow\left( {y,\sigma,c_{coarse}} \right) \right.
\end{equation}

\begin{equation}
c_{fine} = y \times c_{mid} + \left( {1 - y} \right) \times c_{coarse}
\end{equation}

$y$ is generated by spatial MLP, which adds supervision from $c_{mid}$ to spatial MLP during backpropagation, increasing the network's perception angle for estimating concentration $\sigma$. Instead of directly filtering the module's estimation of concentration $\sigma$ on the ray, this ensures the speed of model training and rendering, and allows for adaptation to a wider range of reconstruction scenes.

\subsection{Multiple Performance Evaluation}
NeRF evaluates the rendering effects by calculating the MSE between the voxel rendering result and the Ground Truth. Following volume rendering, Enhance-NeRF obtains three different colors, each representing a distinct meaning. Setting up a ground truth to evaluate different colors is a prerequisite for the model to achieve good results. Spherical harmonic functions are widely used in storing lighting conditions in CG rendering, where only the coefficient expressions of the basic functions need to be stored to represent the image. At any spatial coordinate point, let the base functions of spherical harmonics be:

\begin{equation}
Y_{lm}\left( {\theta,\phi} \right) = \sqrt{\frac{2l + 1}{4\pi} \cdot \frac{\left( {l - m} \right)!}{\left( {l + m} \right)!}}P_{l}^{m}\left( {co{s\theta}} \right)e^{im\phi}
\end{equation}

\begin{equation}
\mathbf{y}\left( \mathbf{d} \right) = \left\lbrack {Y_{0}^{0}\left( \mathbf{d} \right),Y_{1}^{- 1}\left( \mathbf{d} \right),Y_{1}^{0}\left( \mathbf{d} \right),Y_{1}^{1}\left( \mathbf{d} \right),\ldots,Y_{\mathcal{\ell}}^{\mathcal{\ell}}\left( \mathbf{d} \right)} \right\rbrack\mathbf{~}
\end{equation}

In BRDF, given an input angle $\mathbf{d}$ and the corresponding three-channel color $\mathbf{c}(\mathbf{d})$, we can find $\mathbf{k}$ such that $\mathbf{c}(\mathbf{d})=\mathbf{y}(\mathbf{d})\mathbf{k}$. Thus, by adjusting the order $\ell$ (the degree of spherical harmonic functions), we can change the number of coefficients to obtain different degrees of Ground Truth. Through the application of spherical harmonic functions, different levels of ground truth can be obtained as indicated in \textbf{Equation (11)}.

\begin{equation}
C^{'}\left( \mathbf{d} \right) = {\sum\limits_{i = 1}^{\mathcal{\ell}_{max}}{Y_{i}\left( \mathbf{d} \right)\mathbf{k}_{\mathbf{i}}}}
\end{equation}

However, this method requires solving for $\mathbf{k}$ or each sampling point, making it difficult to calculate the loss value. The efficiency of training will be greatly reduced. We propose another method to construct the true values of colors at different levels. By applying spherical harmonic encoding directly to RGB colors, as demonstrated in \textbf{Equation (12)}, we can define the Ground Truth of multiple colors.

\begin{equation}
C^{'}(c) = \left\lbrack {Y_{0}^{0}(c),Y_{1}^{- 1}(c),Y_{1}^{0}(c),Y_{1}^{1}(c),\ldots,Y_{\mathcal{\ell}}^{\mathcal{\ell}}(c)} \right\rbrack
\end{equation}

At the same time, the original size of the RGB color space is $255^3$. However, with Equation 13, we expand it to $\ell^2\times255^3$, which enhances the richness of color signals and enables us to adjust $\ell$ to establish a graded color metric in Enhance-NeRF. This metric ensures that colors undergo varying degrees of transformation while maintaining the same properties. Evaluating scene colors and other factors with continuity may lose the understanding of information such as light and texture when using low-dimensional evaluation standards. In this paper, we set the values $\ell$ for  $c_{fine}$, $c_{mid}$, $c_{coarse}$ to be 4, 3, and 2 respectively. By increasing the weighting of $c_{mid}$, we enhance the model’s sensitivity to lighting information. Therefore, the total loss function of the model is set as follows:

\begin{equation}
\mathcal{L} = \mathcal{L}_{prop} + {\left\| {c -} \right.\left. {~c}_{fine} \right\|}_{2}^{2} + {\sum\limits_{i = 2}^{4}{\left\| {{SH(c)}^{i} -} \right.\left. {SH\left( c_{*} \right)}^{i} \right\|}_{2}^{2}}
\end{equation}

\section{Experimental Results}
\subsection{Datasets}
\noindent \textbf{\textit{Tanks and Temples. }}  Tanks and Temples dataset \cite{knapitsch2017tanks}is used to evaluate our method. We selected scenes, which are outdoor scenes or scenes with significant light reflections. Each scene contains approximately 300 images. 

\begin{figure*}[htbp]
    \centering
    \includegraphics[width=0.95\textwidth]{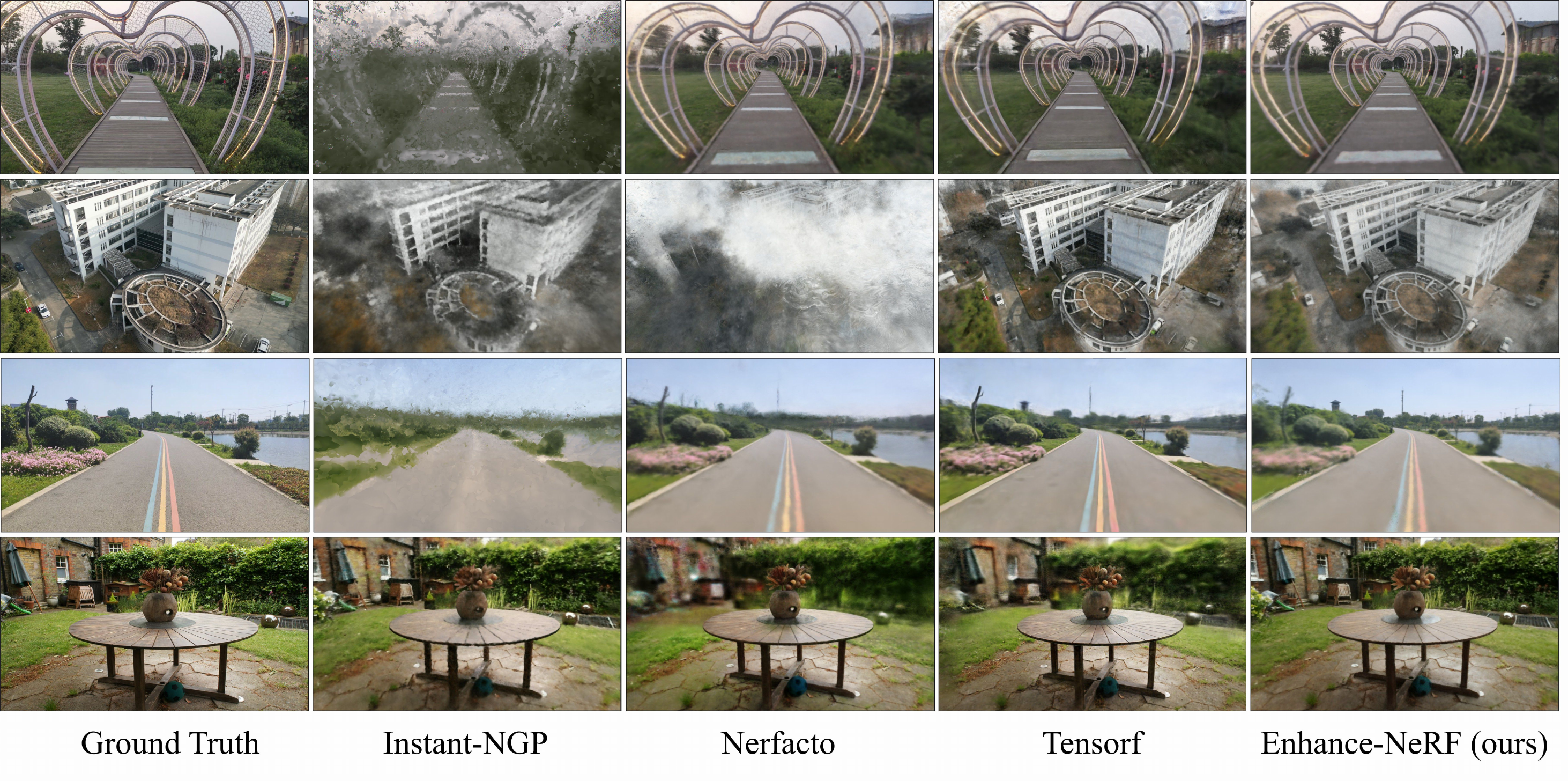}
    \caption{\textbf{Testing of Mixed Sampling dataset. } Enhance-NeRF can adapt to various image acquisition methods and has good texture reconstruction for both distant and close-up scenes, while also removing the "fog" noise in the scene.}
\end{figure*}

\begin{figure*}[htbp]
    \centering
    \includegraphics[width=0.95\textwidth]{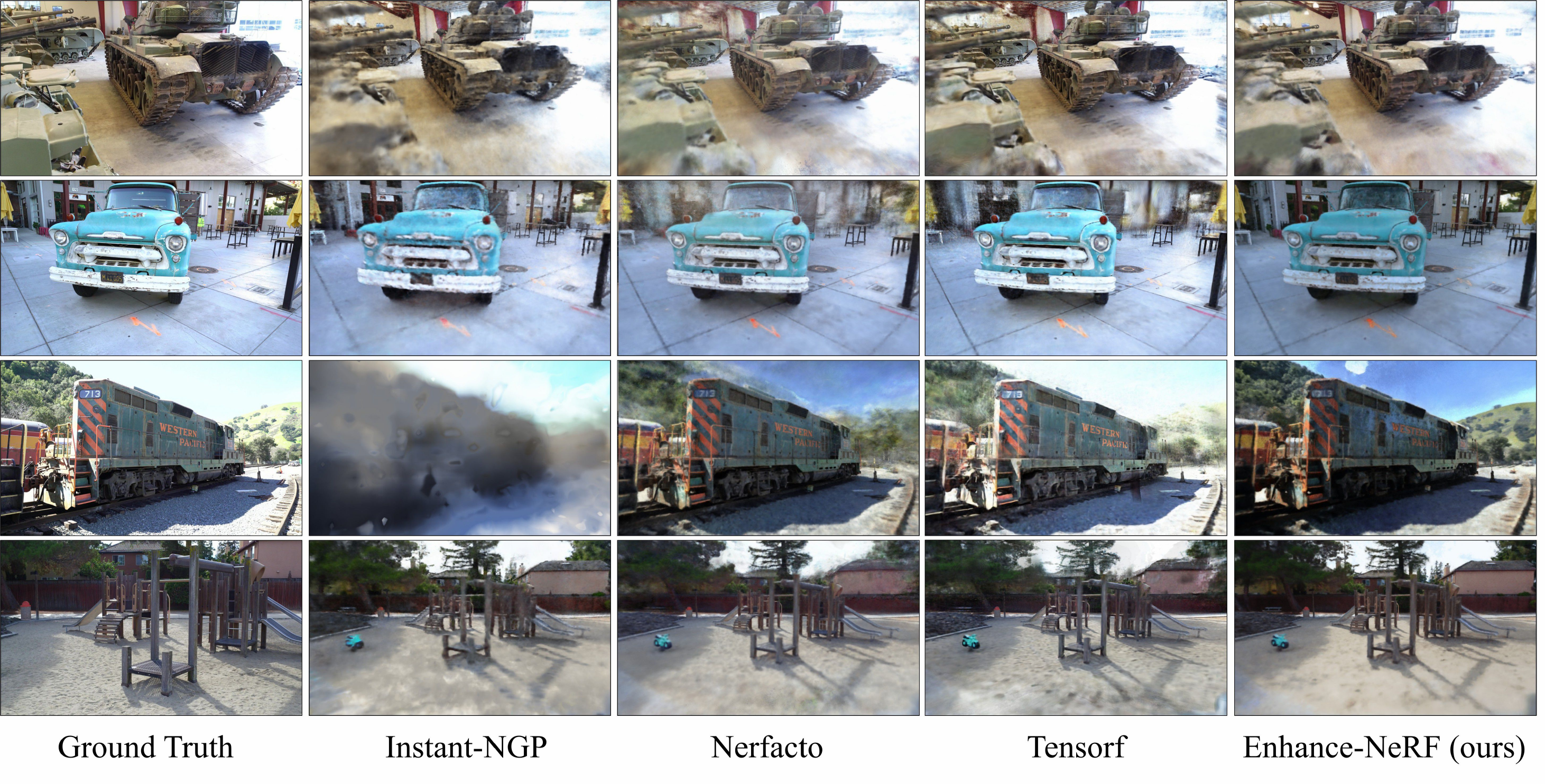}
    \caption{\textbf{Testing of four input forms in the Tanks and Temples dataset.}  All scenes exhibit severe overexposure. When there is intense lighting, Enhance-NeRF can depict high-brightness areas more smoothly and naturally with appropriate adjectives.}
\end{figure*}

\noindent \textbf{\textit{Mixed Sampling. }} The use of a single camera for scene collection often results in uneven sampling of different regions. To better match real shooting habits, we selected scenes of different input forms to create the dataset. The images were captured by shooting a video using DJI Mini 3 Pro drone or mobile phone, and then extracting photos from it. The shooting modes include moving straight forward, self-rotating camera during straight forward movement, circling around the target at the same height (not 360 degrees), and shooting with a fixed angle tilt using a drone, among other relatively arbitrary methods. Each dataset includes 200 to 300 images as input, with missing perspectives that are more consistent with the actual shooting habits of robots and pose some difficulties for the model's reconstruction. And we also included the garden dataset from Mip-NeRF\cite{barron2021mip} and the poster dataset from Nerfacto into it.

\subsection{Optimization and Evaluation}
In this part, we show the detail of our model. We use a spatial MLP with 3 layers and 64 hidden units and a direction MLP with 2 layers and 32 hidden units, both of which use ReLU internal activations. The pre-trained decoder is same with NICE-SLAM . To better demonstrate the model's scene fitting ability, we trained all models and scenes in this paper for only 30,000 iterations, with the number of rays per batch during training set at 4096 on a single RTX 3090 GPU.
To demonstrate the improvements of each module on the model, three error metrics were reported including PSNR, SSIM, and LPIPS \cite{zhang2018unreasonable}. The original sizes of images differ across different scenes, but all rendered output images are standardized to 2048x1080px. \textbf{\textit{Therefore, the actual captured and generated images may not have an identical field of view width.}}

\subsection{Compared methods}
This chapter compares Enhance-NeRF with other models and analyzes the functions of each module. Vanilla NeRF \cite{martin2021nerf}, as the most basic model, has almost no ability to model scenes with a certain scale. Mip-NeRF \cite{barron2021mip} replaces the ray-sampling method used in Vanilla NeRF with an anti-aliasing frustum rendering approach. Instant-NGP \cite{muller2022instant} proposes a multi-resolution hash encoding structure with learnable parameters, instead of the trigonometric frequency encoding used in Vanilla NeRF. Nerfacto \cite{tancik2023nerfstudio} combines the hash encoding of Instant with the scene convergence of Mip-NeRF360. Tensorf \cite{jin2023tensoir} proposed a method to treat a feature grid as a 4-dimensional tensor, and employed traditional tensor decomposition algorithms in radiation field modeling to reduce the dimensionality of high-dimensional data, saving space and compressing data during modeling.

\begin{figure*}[htbp]
    \centering
    \includegraphics[width=0.95\textwidth]{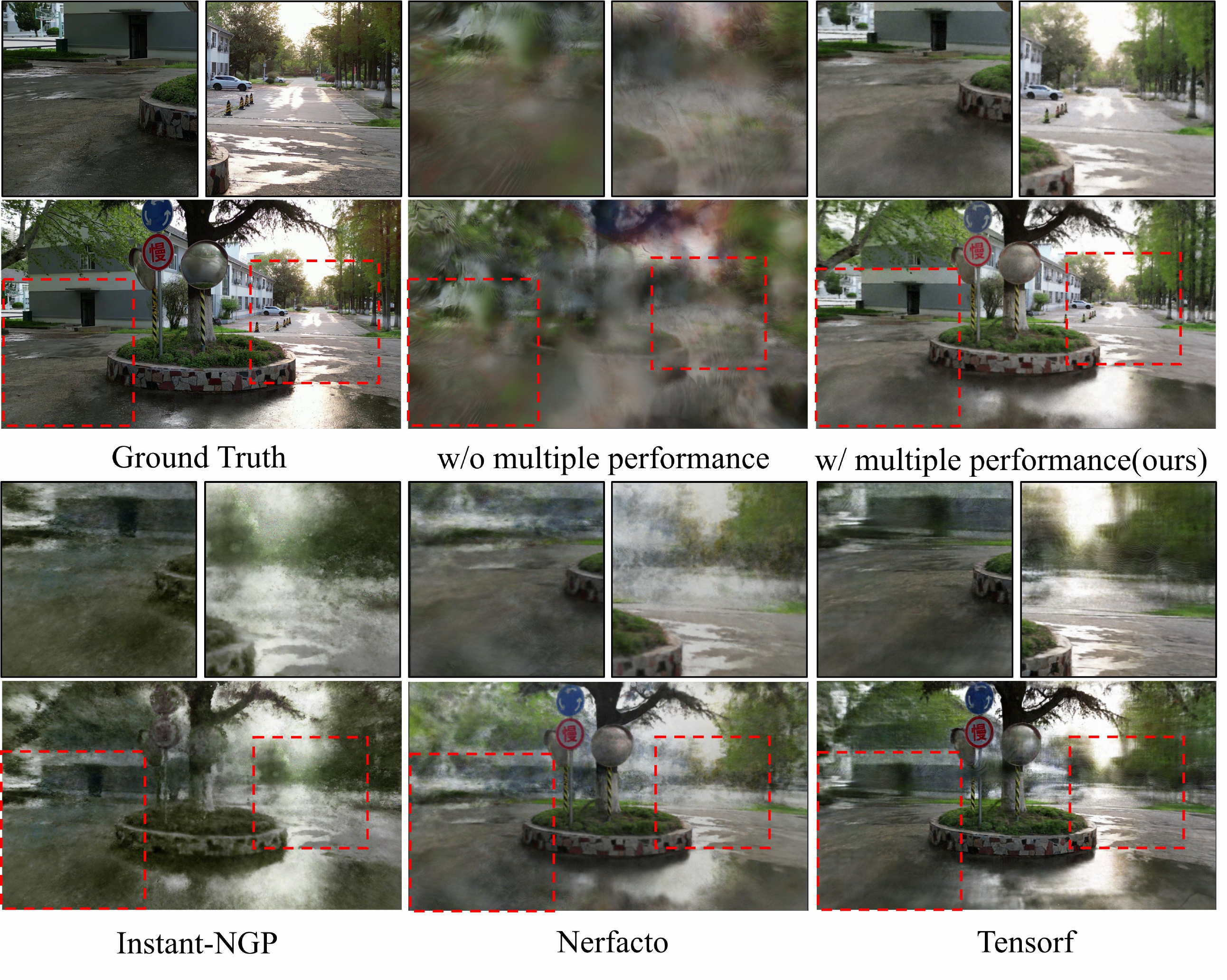}
    \caption{\textbf{Qualitative evaluation on multiple performance.  }The reflective areas in this scene are relatively large, and due to the lack of the multiple performance mechanism, incorrect light reflection processes cannot be suppressed. As a result, the entire scene without multiple performance is enveloped in colored fog. At the same time, in the view with multiple performance mechanisms, the light intensity of different areas within the same viewpoint is adjusted to be nearly consistent.}
\end{figure*}

\begin{figure*}[htbp]
    \centering
    \includegraphics[width=0.95\textwidth]{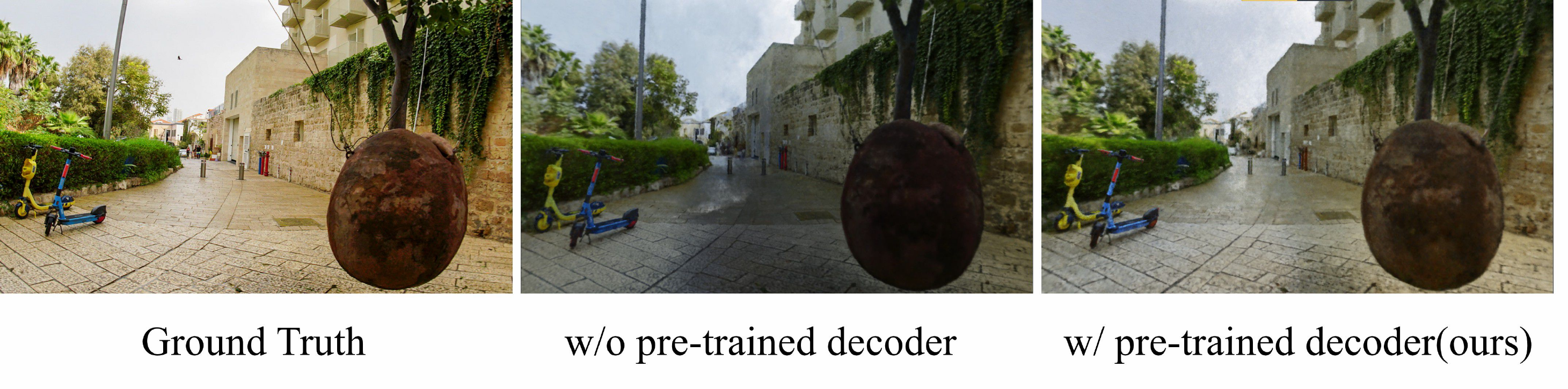}
    \caption{\textbf{Effects of ray perception and reflection-based reconstruction}}
\end{figure*}

\begin{figure*}[htbp]
    \centering
    \includegraphics[width=0.95\textwidth]{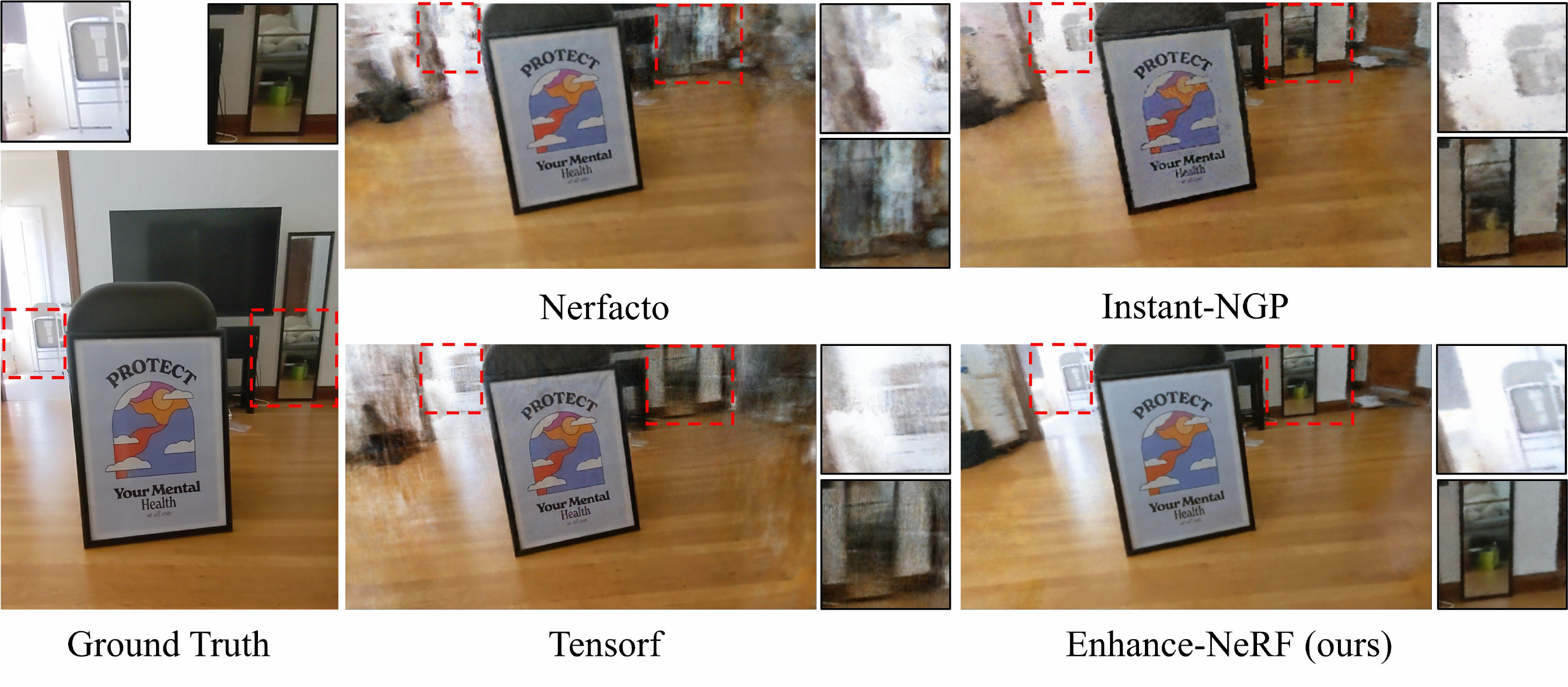}
    \caption{\textbf{ Qualitative evaluation on pre-trained decoder. }Without pre-trained decoder, the stability of the model is weakened, and the success rate of training is reduced. Sometimes, the representation of the scenes is no longer accurate or realistic.}
\end{figure*}

\noindent \textbf{\textit{Appearance in different scene capture.  }}We conducted a comparison of different models using the mixed sampling dataset, presented in \textbf{Fig. 3} and \textbf{Tab. 1}. Vanilla NeRF  and Mip-NeRF  are unable to fit scenes even after being trained for 30,000 iterations. Instant-NGP, which represents scenes using a feature grid, suffers from severe "fog" noise issues in all scenes and loses its scene representation capabilities when presented with forward-style scene input. Tensorf  performs well in all scenes but falls short in reflecting objects in mirrors and displaying distant details. Nerfacto converges quickly in all scenes but lacks good performance in areas beyond the center of the scene and at image edges. Enahnce-NeRF can handle inputs from various scene capture methods, and it performs particularly well in reconstructing the reflection phenomena on object surfaces.

\begin{table}[htbp]
    \centering
    \begin{tabular}{cccrrr}
        \toprule
        Method & & PSNR↑& SSIM↑& LPIPS↓ \\
        \midrule
        Vanilla NeRF && 11.69 & 0.29 & 1.07 \\
        Mip-NeRF && 11.55 & 0.28 & 1.00 \\
        Instant-NGP && 20.42 & 0.51 & 0.60 \\
        Nerfacto && 17.88 & 0.53 & 0.57 \\
        Tensorf && 20.79 & 0.59 & 0.45 \\
        Enhance-NeRF && \textbf{21.17} & \textbf{0.61} &\textbf{ 0.44} \\
        \bottomrule
        \end{tabular}
    \caption{\textbf{Quantitative comparison and evaluation on our Mixed Sampling dataset\textit{. }}We compare our proposed Enhance-NeRF with representative NeRF-based methods and their variants, using datasets captured under different scene capture methods.}
    \label{tab1}
\end{table}

\noindent \textbf{\textit{Appearance in Tanks and Temples dataset. }}In outdoor scenes, overexposure caused by intense one-sided lighting often causes blurred object boundaries. Therefore, the evaluation of the scene reconstruction effect should not only focus on the similarity with the original image, but also pay more attention to surface reconstruction of the objects and reduce emphasis on light propagation. As shown in \textbf{Fig. 4} and \textbf{Tab. 2}, our method can represent the surface materials of the objects well. Although the intensity of light in the overexposed areas has been adjusted during reconstruction, it is represented softly and suits human visual perception better.

\begin{table}[htbp]
    \centering
        \begin{tabular}{cccrrr}
        \toprule
        Method & & PSNR↑ & SSIM↑ & LPIPS↓ \\
        \midrule
        Instant-NGP && 19.59 & 0.57 & 0.54 \\
        Nerfacto && 17.47 & 0.53 & 0.49 \\
        Tensorf && 18.98 & 0.60 & 0.41 \\
        Enhance-NeRF && \textbf{19.82} & \textbf{0.62} & \textbf{0.39} \\
        \bottomrule
        \end{tabular}
    \caption{\textbf{Quantitative comparison and evaluation on Tanks and Temples dataset.}}
    \label{tab2}
\end{table}

\noindent \textbf{\textit{Ablation Study on our Method Design. }}We tested the main components of Enhance-NeRF. Fig. 5 illustrates the performance of the multiple performance mechanisms in complex light reflection scenes, which balance the lighting performance in the scene and keep the light intensity almost consistent. In \textbf{Fig. 6}, without a pre-trained decoder, the reconstruction of light conditions in the environment becomes unstable and black shadows appear at the center of the viewpoint due to the light adjustment caused by the multiple performance mechanisms. The quality evaluations are presented in \textbf{Tab. 3}.

\begin{table}[htbp]
    \centering
        \begin{tabular}{ccccccr}
        \toprule
        Scene & \multicolumn{3}{c}{Tree (Fig.5)} & \multicolumn{3}{c}{Floating-tree (Fig.6)}  \\
        \midrule
        Method & PSNR& SSIM& LPIPS & PSNR& SSIM& LPIPS\\
        \midrule
        w/o Mul-Perf. & 14.28 & 0.32 & 0.82 & 18.53 & 0.47 & 0.58 \\
        w/o Pre-Wei. & 20.72 & 0.46 & 0.58 & 16.70 & \textbf{0.63} & \textbf{0.29} \\
        Enhance-NeRF & \textbf{22.50} & \textbf{0.60} & \textbf{0.38} & \textbf{19.12} & 0.61 & 0.31 \\
        \bottomrule
        \end{tabular}
    \caption{\textbf{Ablation experiment. }By utilizing both multiple performance metrics and pre-trained decoder, Enhance-NeRF can achieve stable performance in various scenes. }
    \label{tab3}
\end{table}

\begin{figure*}[htbp]
    \centering
    \includegraphics[width=0.95\textwidth]{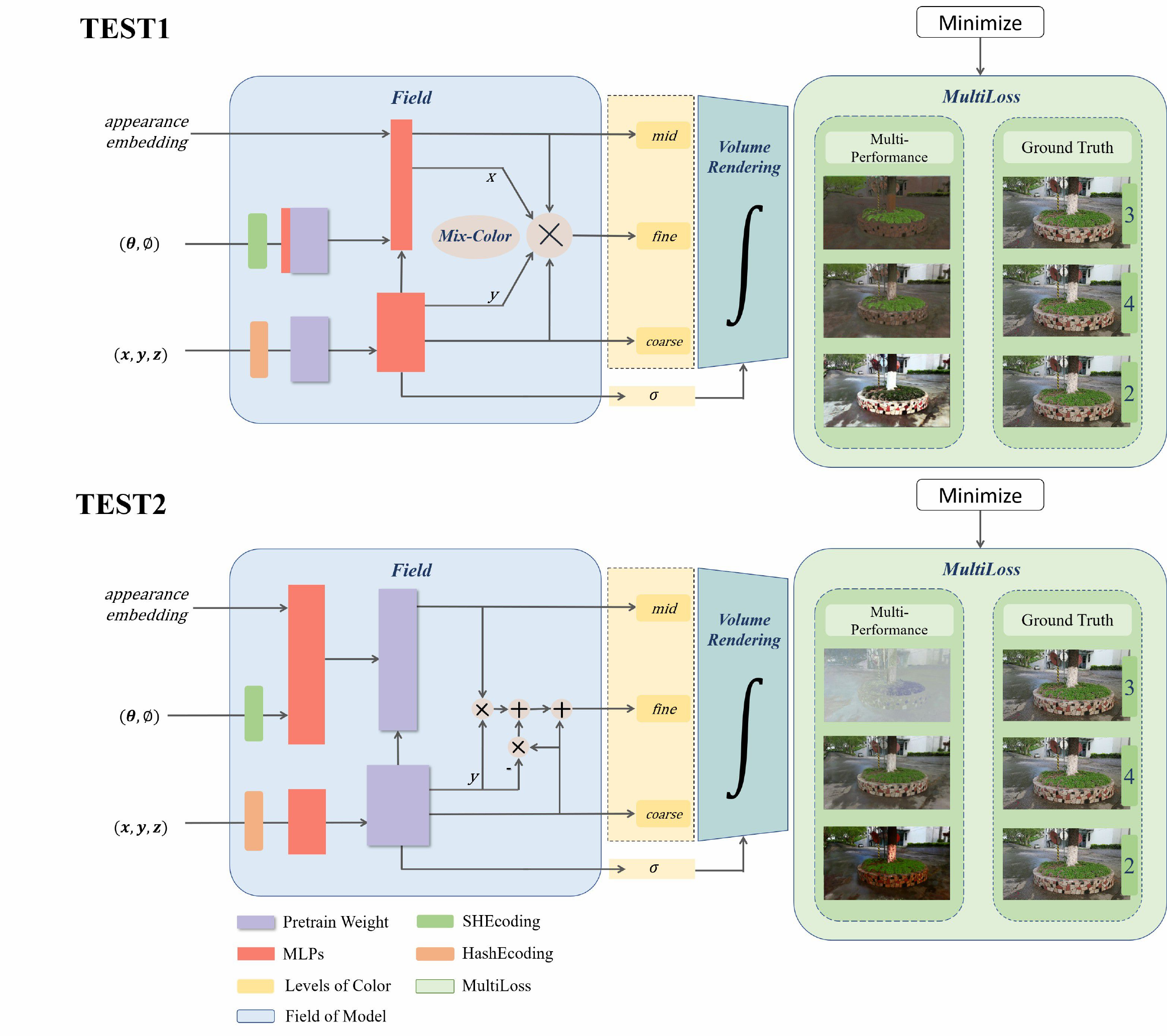}
    \caption{\textbf{ Tricks in model architectures. }Both methods have poor fitting performance on $c_{mid}$.}
\end{figure*}

\section{Discussion and Limitations}

\subsection{Multiple Performance Metrics }
Enhance-NeRF is aimed at outdoor scene reconstruction. By incorporating Multiple Performance Metrics, the scene is divided into three parts based on the reflection requirements of the ray: the space outside the objects, non-reflection areas, and reflection areas on object surfaces. To eliminate the 'fog' noise in the scene, we assume that the scene is in a vacuum state and there is no light reflection caused by dust in the air. Light reflection only occurs on the object surface, therefore, the color of photons outside of objects should not be affected by view-dependent component. The density $\sigma$ of the original NeRF model depends solely on the position of photons, so when the scene background is dense and the reflection of light is complex, there exist biases in estimating the positions of object surfaces. This results in the colored fog seen in \textbf{Fig. 5}. In the sampling process of NeRF, the positions where density $\sigma$ undergoes huge changes come from the surface of objects, and likewise, the positions with maximum color variations of photons on the same ray also come from the surface of objects. The similarity of their distributions enables the distribution of density sigma to be effectively improved when it is supervised by color. However, this color should come from the same position as density in the model (spatial MLP), in order to supervise the distribution of density on the sample ray.
\par
By utilizing volume rendering during changes in viewing angle, “distinct shadows" can be obtained on the surface by combining internal photons of the object, we called middle color. Conversely, for non-reflective areas on the surface, the combination of internal photons of the object will produce the ”same shadow" when the viewing angle is changed, and we call it coarse color, which only depends on its position. We decompose the internal photons of the two regions on the surface of the object according to their functions by changing the proportion of coarse color and middle color.
\par
Considering the two issues collectively, we have increased the output dimensions of the spatial MLP and combined it with the output of the directional MLP. Even though utilizing the incident direction directly for reconstruction, instead of using the reflection direction as in Ref-Nerf, NeRF can still be regarded as a system that contains BDRF functionality. After suppressing the fog problem, mirror reflections in the scene can also be reconstructed well without other complex transformations. As a result, our model can cleanly represent specular reflections in the scene and objects under intense illumination within an hour, as shown in \textbf{Fig. 7}.

\subsection{Pre-trained Decoder}
Inspired by Nice-SLAM, we introduce the pre-trained decoder to enforce fixed information, which increases the fitting difficulty of the model. After introducing Multiple Performance Metrics, Enhance-NeRF will adjust the lighting performance in the scene to maintain consistent illumination intensity across different regions in the image. This alters the lighting behavior of the scene. In \textbf{Fig. 5}, our method achieves consistent illumination intensity between the shadowed area on the left side of the house and the direct sunlight on the right side. Similarly, in \textbf{Fig. 4}, the truck and train show the same effect. 

\begin{figure*}[htbp]
    \centering
    \includegraphics[width=0.95\textwidth]{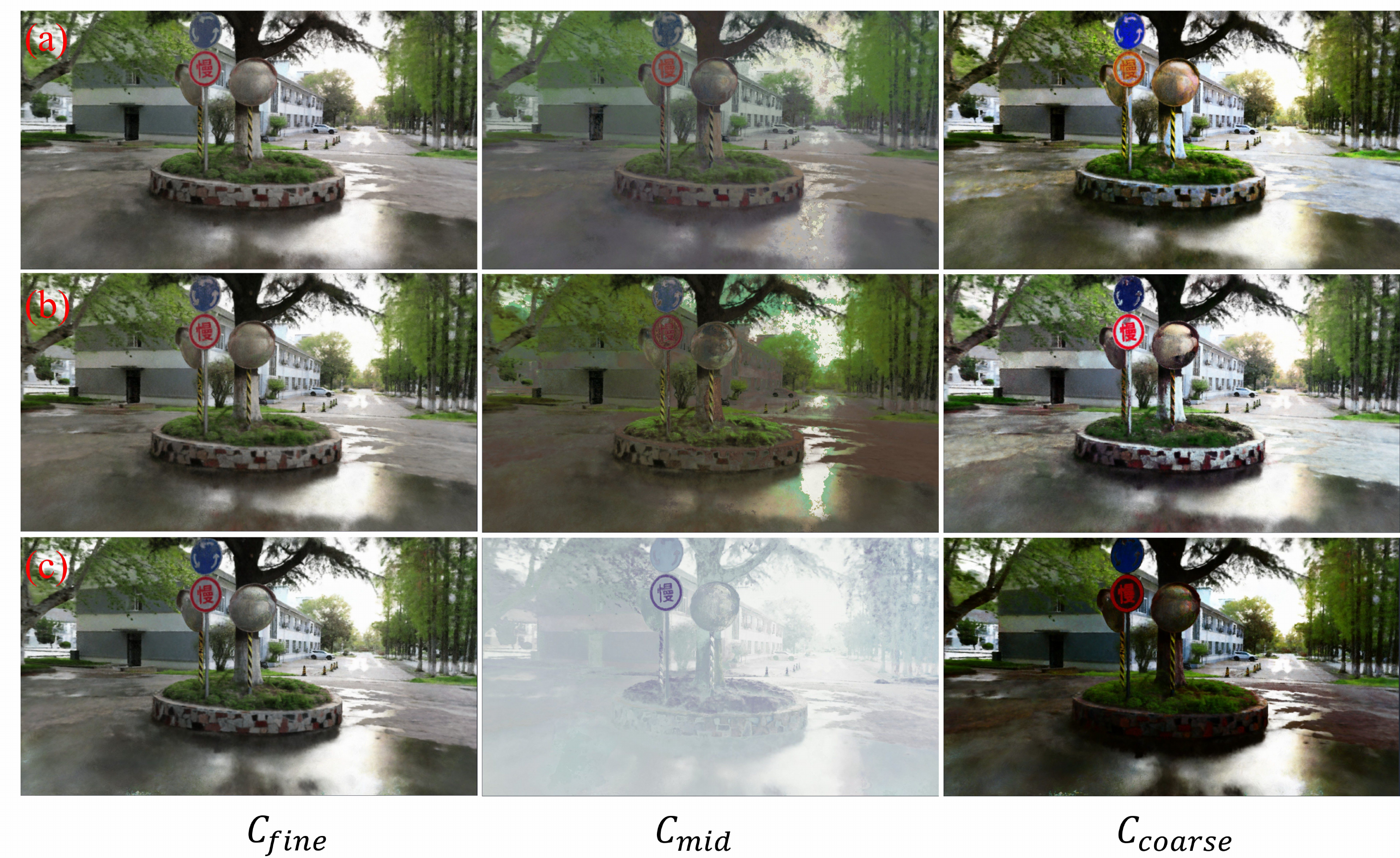}
    \caption{\textbf{Composition of joint color. }From top to bottom, the outputs correspond to Enhance-NeRF, TEST1, and TEST2, respectively.}
\end{figure*}

\par
But without Pre-trained Decoder, Enhance-NeRF often excessively modifies the lighting intensity of the scene, causing black areas to appear in the image as shown in \textbf{Fig. 6}. We believe that during the updating process of spatial MLP using color supervision, there was not a significant difference formed between the direction and position of each sample point, and the global updating mechanism of MLP has affected the established lighting strength of the scene. To achieve rapid fitting of the model and reduce the influence of color during model updating, we introduced the frozen Pre-trained Decoder after encoding the position and direction. This helped to increase the dimensions of the pose encoding and reduce the global impact of color during backpropagation.  
\par
\subsection{Model }
Regarding tricks for adjusting the model, we have made another attempt as shown in \textbf{Fig. 8} and \textbf{Fig. 9}. In TEXT1, we added an additional tuning factor to symmetrize the model outputs and balance the proportion of three colors. The architecture of the model is shown as follows:
\begin{equation}
\left. F_{\Theta 1}:\left( {\mathbf{d},\mathbf{a}} \right)\rightarrow\left( {c_{mid},x} \right) \right.
\end{equation}

\begin{equation}
\left. F_{\Theta 2}:\left( \mathbf{x} \right)\rightarrow\left( {y,\sigma,c_{coarse}} \right) \right.
\end{equation}

\begin{equation}
fine1 = y \times c_{mid} + \left( {1 - y} \right) \times c_{coarse}
\end{equation}

\begin{equation}
fine2 = \left( {1 - x} \right) \times c_{mid} + x \times c_{coarse}
\end{equation}

\begin{equation}
c_{fine} = \frac{1}{2}\left( {fine1 + fine2} \right)
\end{equation}

\par
The first major issue with designing the architecture in this way is establishing the ground truth for the three colors. The purpose of this paper is to improve the clarity of outdoor scene reconstruction and reconstruct light reflections in outdoor scenes. Therefore, the weight of $c_{mid}$ should be higher than that of $c_{coarse}$, which contradicts the setting in TEST1 where the three colors are averaged. TEST2 mimics the architecture of NICE-SLAM, placing the pre-trained decoders at the back end of the model, with MLPs serving as feature storage modules. This setting enables a good reconstruction of the shapes of objects in the scene, but the reproduction of high-level features such as texture and light reflection is not ideal.
\par

\begin{table}[htbp]
    \centering
        \begin{tabular}{cccc}
        \toprule
        Metrics & Enhance-NeRF & TEST1 & TEST2 \\
        \midrule
        PSNR (fine) & \textbf{22.50} & 19.71 & 20.52 \\
        SSIM (fine) & \textbf{0.60} & 0.53 & 0.45 \\
        LPIPS (fine) & \textbf{0.38} & 0.44 & 0.59 \\
        \midrule
        PSNR (mid) & \textbf{18.26} & 13.66 & 7.94 \\
        SSIM (mid) & \textbf{0.48} & 0.40 & 0.28 \\
        LPIPS (mid) & \textbf{0.49} & 0.58 & 0.80 \\
        \midrule
        PSNR (coarse) & \textbf{17.80} & 16.51 & 14.60 \\
        SSIM (coarse) & \textbf{0.55} & 0.55 & 0.34 \\
        LPIPS (coarse) &\textbf{0.42 }& 0.42 & 0.64 \\
        \bottomrule
        \end{tabular}
    \caption{\textbf{Performance evaluation of three-channel color. }}
    \label{tab4}
\end{table}

\par
Illustrated in \textbf{Tab. 4}, Enhance-NeRF achieved the best results in all metrics for images with three channels, which demonstrates the effectiveness of our loss function design strategy. Perhaps with increased batch size and training time, the performance of the three methods may become more consistent, and the performance of Enhance-NeRF can be further improved.

\subsection{Limitations }
Although Enhance-NeRF can achieve good results in various scenes, it can effectively remove the haze noise in the scene. However, there are still some issues to be addressed. It lacks detail representation of faraway objects in unbounded scenes. Meanwhile, due to the lack of improvement on the image capture scheme, all models have a limited perspective during the reconstruction process. When the clarity of mirrored reflections in the image is insufficient, the rendering of the model lacks sufficient expressive power. Additionally, since Enhance-NeRF employs PyTorch for model construction, there is still room for efficiency improvement. Methods such as embedding MLP head into directional MLP and spatial MLP and using tiny-cudnn for model building can be employed to enhance the method's efficiency.

\section{Conclusion}
This paper proposes Enhance-NeRF, a pure MLP-based NeRF, for fast fitting of outdoor scenes, by combining several papers' components and introducing Multiple Performance Metrics, Pre-trained Decoder. Our experiments show that our method can represent outdoor unbounded scenes more cleanly compared to other fast-converging models. Coeds and dataset will be released.

\section{Acknowledgements}
This work was supported by the Jiangsu Modern Agricultural Equipment and Technology Demonstration \& Promotion Project (project No.NJ2021-60), and the Jiangsu Agricultural Science and Technology Innovation Fund (JASTIF)(Grant no.CX (21) 3146).

\printbibliography

\end{document}